%% file: 121-main.tex
\documentclass[runningheads]{llncs}

\usepackage{graphicx}
\usepackage{comment}
\usepackage{amsmath,amssymb}
\usepackage{color}
\usepackage{url}
\usepackage{hyperref}

\newif\ifreview
\reviewfalse

\ifreview
	\usepackage{lineno}

	\linenumbers
\fi

\usepackage{booktabs}
\usepackage{subcaption}
\usepackage{multirow}
\usepackage{algorithm,algorithmic}
\usepackage{mathtools}

\begin{document}

%%%%%%%%%%%%%%%%%%%%% Add submission id, track, and title. %%%%%%%%%%%%%%%%%%%%%

% Insert your submission number here
\def\SubNumber{121}

% Choose one track by uncommenting one of the following lines  
\def\GCPRTrack{Regular Track}
% \def\GCPRTrack{Track: Computer vision systems and applications}
% \def\GCPRTrack{Track: Pattern recognition in the life and natural sciences}
% \def\GCPRTrack{Track: Photogrammetry and remote sensing}
% \def\GCPRTrack{Track: Robot vision}
% \def\GCPRTrack{Track: DAGM Young Researcher Forum}

% Replace with your title
\title{Learning Conditional Invariance through Cycle Consistency}
% You can use \thanks for acknowledgment. Do not add any acknowledgment to the draft 
% version that is used for the review process.  
%\title{Title\thanks{XXX}}

\ifreview
	% ANONYMOUS SUBMISSION FOR REVIEW
	% DO NOT MODIFY these for the draft version that is used for the review process.
	\titlerunning{DAGM GCPR 2021 Submission \SubNumber{}. CONFIDENTIAL REVIEW COPY.}
	\authorrunning{DAGM GCPR 2021 Submission \SubNumber{}. CONFIDENTIAL REVIEW COPY.}
	\author{DAGM GCPR 2021 - \GCPRTrack{}}
	\institute{Paper ID \SubNumber}
\else

	\author{Maxim Samarin\inst{1,}\thanks{Both authors contributed equally.}
% 	\orcidID{0000-0002-9242-1827} 
	\and
	Vitali Nesterov\inst{1,\star}
% 	\orcidID{0000-0003-4133-1079} 
	\and
	Mario Wieser\inst{1}
% 	\orcidID{0000-0002-8737-3605} 
	\and
	Aleksander Wieczorek\inst{1}
% 	\orcidID{0000-0003-1565-1781} 
	\and 
	Sonali Parbhoo \inst{2}
% 	\orcidID{0000-0001-8400-3732} 
	\and 
	Volker Roth\inst{1}
% 	\orcidID{0000-0003-0991-0273}
	}
	
	\authorrunning{M. Samarin et al.}
	
	\institute{
	Department of Mathematics and Computer Science, University of Basel, Spiegelgasse 1, 4051 Basel, Switzerland \\
	\email{\{\href{mailto:maxim.samarin@unibas.ch}{maxim.samarin}, \href{mailto:vitali.nesterov@unibas.ch}{vitali.nesterov}, \href{mailto:aleksander.wieczorek@unibas.ch}{aleksander.wieczorek},\\ \href{mailto:volker.roth@unibas.ch}{volker.roth}\}@unibas.ch}, \email{\href{mailto:wieser.mario@gmail.com}{wieser.mario@gmail.com}}
	\and
% 	Genedata AG, Margarethenstrasse 38, 4053 Basel, Switzerland \\
% 	\email{\href{mailto:wieser.mario@gmail.com}{wieser.mario@gmail.com}}
% 	\and
	Harvard John A. Paulson School of Engineering and Applied Sciences,\\Harvard University, 150 Western Ave, Boston, MA 02134, USA \\
	\email{\href{mailto:sparbhoo@seas.harvard.edu}{sparbhoo@seas.harvard.edu}}
	}
\fi

\maketitle              % typeset the header of the contribution
\begin{abstract}
Identifying meaningful and independent factors of variation in a dataset is a challenging learning task frequently addressed by means of deep latent variable models. This task can be viewed as learning symmetry transformations preserving the value of a chosen property along latent dimensions. However, existing approaches exhibit severe drawbacks in enforcing the invariance property in the latent space. We address these shortcomings with a novel approach to cycle consistency. Our method involves two separate latent subspaces for the target property and the remaining input information, respectively. In order to enforce invariance as well as sparsity in the latent space, we incorporate semantic knowledge by using cycle consistency constraints relying on property side information. The proposed method is based on the deep information bottleneck and, in contrast to other approaches, allows using continuous target properties and provides inherent model selection capabilities. We demonstrate on synthetic and molecular data that our approach identifies more meaningful factors which lead to sparser and more interpretable models with improved invariance properties.
\end{abstract}
\input{1-intro}
\input{2-related_work}
\input{3-preliminaries}
\input{4-model}
\input{5-experiments}
\input{6-discussion}
\input{7-acknowledgements.tex}
%
% Add citations of the appendix
\nocite{tensorflow2015-whitepaper}
\nocite{adam}
% \newpage
\bibliographystyle{splncs04}
\bibliography{10-references}
%
% To be split
%\newpage
%\input{11-appendix}
%
\end{document}

%% file: 1-intro.tex
\section{Motivation}
\label{intro}
Understanding the nature of a generative process for observed data typically involves uncovering explanatory factors of variation responsible for the observations. But the relationship between these factors and our observation usually remains unclear. A common assumption is that the relevant factors can be expressed by a low-dimensional latent representation $Z$ \cite{locatello2019challenging}. Therefore, popular machine learning methods involve learning of appropriate latent representations to \textit{disentangle} factors of variation. Learning disentangled representations is often considered in an unsupervised setting which does not rely on the prior knowledge about the data such as labels \cite{chen2018isolating,chen2016infogan,Higgins2017betaVAELB,kim2018factorVAE,lin2020infogan}. However, it was shown that inductive bias on the dataset and learning approach is necessary to obtain disentanglement \cite{locatello2019challenging}. Inductive biases allow us to express assumptions about the generative process and to prioritise different solutions not only in terms of disentanglement \cite{multi_level_vae,Higgins2017betaVAELB,Klys,robert2019dualdis,wieser2020inverse}, but also in terms of constrained latent space structures \cite{keller2020learning,archetypes}, preservation of causal relationships \cite{wieczorek2016causal}, or interpretability \cite{wu2017sparsity}.

We consider a supervised setting where semantic knowledge about the input data allows structuring the latent representation in disjoint subspaces $Z_0$ and $Z_1$ of the latent space $Z$ by enforcing conditional invariance. In such supervised settings, disentanglement can be viewed as an extraction of level sets or symmetries inherent to our data $X$ which leave a specified property $Y$ invariant. An important application in that direction is the generation of diverse molecular structures with similar chemical properties \cite{wieser2020inverse}. The goal is to disentangle factors of variation relevant for the property. Typically, level sets $L_y$ are defined implicitly through $L_y(f)=\{(x_1,...,x_d)  \vert f(x_1,...,x_d)=y\}$ for a property $y$ which implicitly describes the level curve or surface w.r.t. inputs $(x_1,...,x_d)\in\mathbb{R}^d$. The topic of this paper is to identify a sparse parameterisation of level sets which encodes conditional invariances and thus selects a correct model. Several techniques have been developed to steer model selection by sparsifying the number of features, e.g. \cite{tibshirani96regression,art:tishby:ib}, or compressing features into a low-dimensional feature space, e.g. \cite{AlemiFD016,rey14,Wieczorek}. These methods improve generalisation by focusing on only a subset of relevant features and using these to explain a phenomenon. Existing methods for including such prior knowledge in the model usually do not include dimensionality reduction techniques and perform a hand-tuned selection \cite{keller2020learning,Klys,wieser2020inverse}.

In this paper, we introduce a novel approach to cycle consistency, relying on property side information $Y$ as our semantic knowledge, to provide conditional invariance in the latent space. With this we mean that conditioning on part of the latent space, i.e. $Z_0$, allows property-invariant sampling in the latent space $Z_1$. By ensuring that our method consistently performs on generated samples when fed back to the network, we achieve more disentangled and sparser representations. Our work builds on \cite{Wieczorek}, where a general sparsity constraint on latent representations is provided, and on \cite{Klys,wieser2020inverse}, where conditional invariance is obtained through adversarial training. We show that our approach addresses some drawbacks in previous approaches and allows us to identify more meaningful factors for learning better models and achieve improved invariance performance. Our contributions may thus be summarised as follows:
\begin{itemize}
\item We propose a novel approach for supervised disentanglement where conditional invariance is enforced by a novel cycle consistency on property side information. This facilitates the guided exploration of the latent space and improves sampling with a fixed property.\vspace{0.1cm}
\item Our model inherently favours sparse solutions, leading to more interpretable latent dimensions and facilitates built-in model selection.\vspace{0.1cm}
\item We demonstrate that our method improves on the state-of-the-art performance for conditional invariance as compared to existing approaches on both synthetic and molecular benchmark datasets.
\end{itemize}

%% file: 2-related_work.tex
\section{Related Work}
\label{related_work}
\subsection{Deep Generative Latent Variable Models and Disentanglement}
Because of its flexibility, the variational autoencoder (VAE) \cite{kingma2013auto,rezende14} is a popular deep generative latent variable model in many areas such as fairness \cite{louizos2015variational}, causality \cite{Louizos}, semi-supervised learning \cite{NIPS2014_5352}, and design and discovery of novel molecular structures \cite{Gomez,kusner,nesterov20203dmolnet}.
The VAE is closely related to the Information Bottleneck (IB) principle \cite{AlemiFD016,art:tishby:ib}. Various approaches exploit this relation, e.g the deep variational information bottleneck (DVIB) \cite{achille2018information,AlemiFD016}. Further extensions were proposed in the context of causality \cite{chicharro2020causal,parbhoo20a,parbhoo2020information} or archetypal analysis \cite{keller2020learning,archetypes}. 

The $\beta$-VAE \cite{Higgins2017betaVAELB} extends the standard VAE approach and allows unsupervised disentanglement. In unsupervised settings, there exists a great variety of approaches based on VAEs and generative adversarial networks (GANs) to achieve disentanglement such as FactorVAE \cite{kim2018factorVAE}, $\beta$-TCVAE \cite{chen2018isolating} or InfoGAN \cite{chen2016infogan,lin2020infogan}. Partitioning the latent space into subspaces is inspired by the multi-level VAE \cite{multi_level_vae}, where the latent space is decomposed into a local feature space that is only relevant for a subgroup and a global feature space. In supervised settings, several approaches such as \cite{Creswell,Klys,FaderNetworks,wieser2020inverse} achieve disentanglement by applying adversarial information elimination to select a model with partitioned feature and property space. In such a setting, different to unsupervised disentanglement, our goal is supervised disentanglement with respect to a particular target property.

Another important line of research employs the idea of cycle consistency for learning disentangled representations. Presumably the most closely related work to this study is conducted by \cite{Jha2018DisentanglingFO,mariophd,wieser2020inverse}. Here, the authors employ a cycle-consistent loss on the latent representations to learn symmetries and disentangled representations in weakly supervised settings, respectively. Moreover, in \cite{wieser2020inverse}, the authors use adversarial training and mutual information estimation to learn symmetry transformations instead of explicitly modelling them. In contrast, our work replaces adversarial training by using cycle consistency.

\subsection{Model Selection via Sparsity}
Several works perform model selection by introducing sparsity constraints which penalise the model complexity. A common sparsity constraint is the Least Absolute Shrinkage and Selection Operator (LASSO) \cite{tibshirani96regression}. Extensions of the LASSO propose a log-penalty to obtain even sparser solutions in the compressed IB setting \cite{rey14} and generalise it further to deep generative models \cite{Wieczorek}. Furthermore, the LASSO has been extended to the group LASSO, where combinations of covariates are set to zero, the sparse group LASSO \cite{Simon13asparse-group}, and the Bayesian group LASSO \cite{sudhir}. Perhaps most closely related to our work is the oi-VAE \cite{ainsworth18a}, which incorporates a group LASSO prior in deep latent variable models. These methods employ a general sparsity constraint to achieve a sparse representation. Our model extends these ideas and imposes a semantic sparsity constraint in the form of cycle consistency that performs regularisation based on prior knowledge. 

%% file: 3-preliminaries.tex
\section{Preliminaries}
\label{prelim}
\subsection{Deep Variational Information Bottleneck}
We focus on the DVIB \cite{AlemiFD016} which is a method for information compression based on the IB principle \cite{art:tishby:ib}. The objective is to compress a random variable $X$ into a random variable $Z$ while being able to predict a third random variable $Y$.
The DVIB is closely related to the VAE \cite{kingma2013auto,rezende14}. The optimal compression is achieved by solving the parametric problem
\begin{equation}
\min_{\phi,\theta}  I_{\phi}(Z;X) - \lambda  I_{\phi,\theta}(Z;Y),
\label{eq:parametricib}
\end{equation}
where $I$ is the mutual information between two random variables. Hence, the DVIB objective balances maximisation of $I_{\phi,\theta}(Z;Y)$, i.e. $Z$ being informative about $Y$, and minimisation of $I_{\phi}(Z;X)$, i.e. compression of $X$ into $Z$. We assume a parametric form of the conditionals $p_\phi(Z|X)$ and $ p_\theta(Y|Z)$ with $\phi$ and $\theta$ representing the parameters of the encoder and decoder network, respectively.
Parameter $\lambda$ controls the degree of compression and is closely related to $\beta$ in the $\beta$-VAE \cite{Higgins2017betaVAELB}. The relationship to the VAE becomes more apparent with the definition of the mutual information terms:
\begin{align}
I_\phi(Z;X) &= \mathbb{E}_{p(X)}  D_{KL}(p_\phi(Z|X)\| p(Z)) \label{eq:ib:enc},\\
I_{\phi,\theta}(Z;Y) &\geq \mathbb{E}_{p(X,Y)} \mathbb{E}_{p_\phi(Z|X)}\log p_\theta(Y|Z) + h(Y)\label{eq:ib:dec},
\end{align}
with $D_{KL}$ being the Kullback-Leibler divergence, and $h(Y)$ the entropy. Note that we write Eq.~(\ref{eq:ib:dec}) as an inequality which uses the insight of \cite{alex_bound} that the RHS is in fact a lower bound to $I_\theta(Z;Y)$; see \cite{alex_bound} for details.
\subsection{Cycle Consistency}
We use the notion of cycle consistency similar to \cite{Jha2018DisentanglingFO,cyclegan}. The CycleGAN \cite{cyclegan} performs unsupervised image-to-image translation, where a data point is mapped to its initial position after being transferred to a different space. For instance, suppose that domain $X$ consists of summer landscapes, while domain $Y$ consists of winter landscapes (see Appendix Fig. A1). A function $f(x)$ may be used to transform a summer landscape $x$ to a corresponding winter landscape $y$. Similarly, function $g(y)$ maps $y$ back to the domain $X$. The goal of cycle consistency is to learn a mapping to $\hat{x}$, which is close to the initial $x$. In most cases, there is a discrepancy between $x$ and $\hat{x}$ referred to as the cycle consistency loss. In order to obtain an almost invertible mapping, the loss $\lVert (g(f(x)) - x \rVert_1$ is minimised.

%% file: 4-model.tex
\section{Model}
\label{model}
\begin{figure*}[t]
\includegraphics[width=0.889\textwidth]{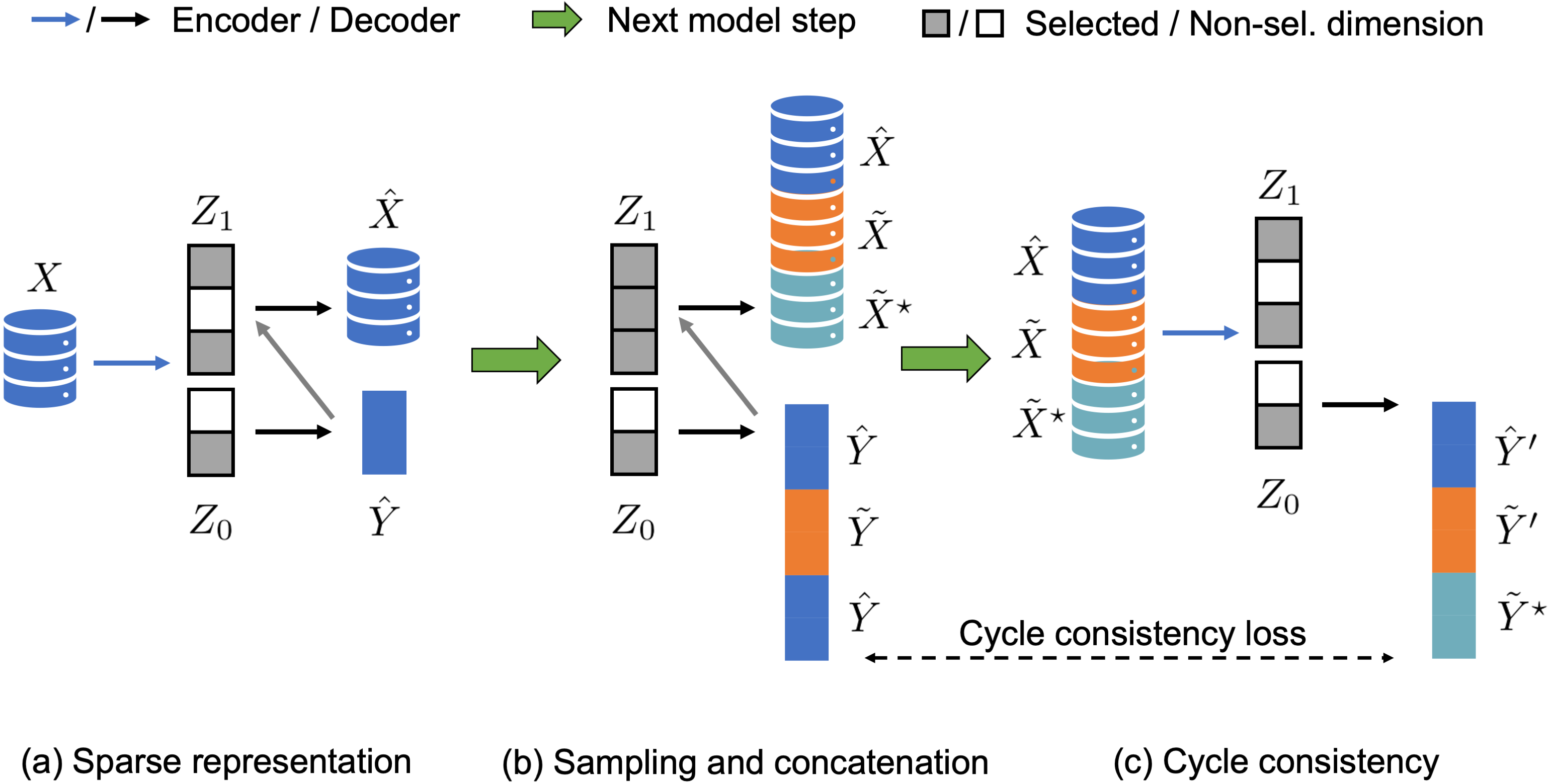}
\centering
\caption{\textbf{Model illustration.} (a) Firstly, we learn a sparse representation $Z$ from our input data $X$ which we separate into a property space $Z_0$ and an invariant space $Z_1$. Given this representation, we try to predict the property $\hat{Y}$ and reconstruct our input $\hat{X}$. Grey arrows indicate that $\hat{Y}=\text{dec}_Y(Z_0)$ instead of $Z_0$ is used for decoding $\hat{X}$ (see Sec. \ref{sec:complete_framework}). (b) Secondly, we sample new data in two ways: (i) uniformly in $Z$ to get new data points $\tilde{X}$ and $\tilde{Y}$ (orange data), (ii) uniformly in $Z_1$ with fixed $Z_0$ to get $\tilde{X}^\star$ at fixed $\hat{Y}$ (cyan data). We concatenate the respective decoder outputs. (c) Lastly, we feed the concatenated input batch $X^\text{c}$ into our model and calculate the cycle consistency loss between the properties.}
\label{fig:completemodel}
\end{figure*}
Our model is based on the DVIB to learn a compact latent representation. The input $X$ and the output $Y$ may be complex objects and take continuous values, such as molecules with their respective molecular properties. Unlike the standard DVIB, we do not only want to predict $Y$ from an input $X$, but also want to generate new $\tilde{X}$ by sampling from our latent representation. As a consequence, we add an additional second decoder that reconstructs $X$ from $Z$ (similar to \cite{Gomez} for decoder $Y$ in the VAE setting), leading to the adjusted parametric objective
\begin{equation}
\min_{\phi,\theta,\tau}  I_{\phi}(Z;X) - \lambda \big( I_{\phi,\theta}(Z;Y) + I_{\phi,\tau}(Z;X)\big),
\label{eq:parametricib:2}
\end{equation}
where $\phi$ are the encoder parameters, and $\theta$ and $\tau$ describe network parameters for decoding $Y$ and $X$, respectively.
\subsection{Learning a Compact Representation}
\label{sec:model_sparseIB}
Formulating our model as a DVIB allows leveraging properties of the mutual information with respect to learning compact latent representations. To see this, first assume that $X$ and $Y$ are jointly Gaussian-distributed which leads to the \textit{Gaussian Information Bottleneck}~\cite{art:chechik:gib} where the solution $Z$ can be found analytically and proved to be Gaussian.
In particular, for $X \sim \mathcal{N}(0, \Sigma_{X})$,
the optimal $Z$ is a noisy projection of $X$: $Z = AX + \xi$, where $\xi \sim \mathcal{N}(0,I)$.
The mutual information between $X$ and $Z$ is then equal to
\begin{equation}
    I(X;Z) = \tfrac{1}{2} \log|A\Sigma_{X}A^\top + I|.
    \label{eq:GIB}
\end{equation}
If we now assume $A$ to be diagonal, the model becomes sparse \cite{rey14}. This is because a full-rank projection $A X^\prime$ of $X^\prime$ does not change the mutual information since $I(X;X^\prime)=I(X;AX^\prime)$. A reduction in mutual information can only be achieved by a rank-deficient matrix $A$. In general, the conditionals $Z|X$ and $Y|Z$ in Eq.~(\ref{eq:parametricib}) may be parameterised by neural networks with $X$ and $Z$ as input. The diagonality constraint on $A$ does not cause any loss of generality of the DVIB solution as long as the neural network encoder $f_\phi$ makes it possible to diagonalise $Af_\phi(X)f_\phi(X)^\top A^\top$ (see \cite{Wieczorek} for more details). In the following, we consider $A$ to be diagonal and define the sparse representation as the dimensions of the latent space $Z$ selected by the non-zero entries of $A$. Recalling Eq. \eqref{eq:GIB}, this allows us to approximate the mutual information for the encoder in Eq. \eqref{eq:ib:enc} in a sparse manner
\begin{equation}
    I_\phi(X;Z) = \tfrac{1}{2}\log|\text{diag} (f_\phi(X)f_\phi(X)^\top) + \mathbf{1}|,
    \label{eq:sparseIB_encoder}
\end{equation}
where $\mathbf{1}$ is the all-one vector and the diagonal elements of $A$ are subsumed in the encoder $f_\phi$.
\subsection{Conditional Invariance and Informed Sparsity}
\label{cycle:model}
A general sparsity constraint is not sufficient to ensure that latent dimensions indeed represent independent factors. In a supervised setting, our target $Y$ conveys semantic knowledge about the input $X$, e.g. a chemical property of a molecule. To incorporate semantic knowledge into our model, we require a mechanism that partitions the representation such that it encodes the semantic meaning not only sparsely but preferably independently of other information concerning the input.

To this end, the central element of our approach is cycle consistency with respect to target property $Y$, which is illustrated in steps (b) and (c) in Fig. \ref{fig:completemodel}. The idea is, that reconstructed $\hat{X}$ or newly sampled $\Tilde{X}$ with associated prediction $\hat{Y}$ and $\Tilde{Y}$ are expected to provide matching predictions $\hat{Y}'$ and $\Tilde{Y}'$ when $\hat{X}$ and $\Tilde{X}$ are used as an input to the network. This means, if we perform another cycle through the network with sampled or reconstructed inputs, the property prediction should stay consistent. 
The partitioning of the latent space $Z$ in the property subspace $Z_0$ and the \textit{invariant} subspace $Z_1$ is crucial. The property $Y$ is predicted from $Z_0$, while the input is reconstructed from the full latent space $Z$. Ensuring cycle consistency with respect to the property allows putting property-relevant information into the property subspace $Z_0$. Furthermore, the latent space is regularised by drawing samples which adhere to cycle consistency and provide additional sparsity. If information about $Y$ is encoded in $Z_1$, this will lead to a higher cycle consistency loss. In this way, cycle consistency enforces invariance in subspace $Z_1$. By fixing coordinates in $Z_0$, and thus fixing a property, sampling in $Z_1$ results in newly generated $\Tilde{X}$ with the same property $\Tilde{Y}$. More formally, fixing $Z_0$ renders random variables $X$ and $Y$ conditionally independent, i.e. $X \perp\!\!\!\perp Y \vert Z_0$ (see Appendix Fig. A2). We ensure conditional invariance with a particular sampling: We fix the $Z_0$ coordinates and sample in $Z_1$ to obtain generated $\Tilde{X}^\star$ all with a fixed property $\hat{Y}$. Using these inputs allows to obtain a new prediction $\Tilde{Y}^\star$ which should be close to the fixed target property $\hat{Y}$. We choose the L$_2$ norm for convenience and define the full cycle consistency loss by
\begin{equation}
    \mathcal{L}_\text{cycle} = 
    \lVert \hat{Y} - \hat{Y}' \rVert _2 
    + 
    \lVert \Tilde{Y} - \Tilde{Y}'\rVert _2  
    + 
    \lVert \hat{Y} - \Tilde{Y}^\star \rVert _2 . 
    \label{eq:cycleib}
\end{equation}
\subsection{Proposed Framework}
\label{sec:complete_framework}
The resulting model in Eq. \eqref{eq:complete} combines sparse DVIBs with partitioned latent space and a novel approach to cycle consistency, which drives conditional invariance and informed sparsity in the latent structure. This allows latent dimensions in $Z_0$ relevant for prediction of $Y$ to disentangle of latent dimensions in $Z_1$ which encode remaining input information of $X$.  
\begin{eqnarray}
  \mathcal{L} &= I_{\phi}(X;Z) - \lambda \Big(&I_{\phi,\tau}(Z_0,Z_1;X) + I_{\phi,\theta}(Z_0;Y) \nonumber \\ 
  &  &-\beta \big( 
    \lVert \hat{Y} - \hat{Y}' \rVert _2 
    + 
    \lVert \Tilde{Y} - \Tilde{Y}'\rVert _2  
    + 
    \lVert \hat{Y} - \Tilde{Y}^\star \rVert _2  
    \big)\Big)
\label{eq:complete}
\end{eqnarray}
The proposed model performs model selection as it inherently favours sparser latent representations. This in turn facilitates easier interpretation of latent factors because of the built-in conditional independence between property space $Z_0$ and invariant space $Z_1$. These adjustments address some of the issues of the STIB \cite{wieser2020inverse} relying on adversarial training, mutual information estimation (which can be difficult in high-dimensions \cite{song2019understanding}) and bijective mapping which can make the training challenging. In contrast to the work of \cite{Jha2018DisentanglingFO}, we impose a novel cycle consistency loss on the predicted outputs $Y$ instead of the latent representation $Z$. A reason to consider rather $Y$ than $Z$ is that varying latent dimensionality leads to severe problems in the optimisation process as it requires an adaptive rescaling of the different loss weights. To overcome this drawback, we close the full cycle and define the loss on the outputs. Appendix Sec. A.3 and Algorithm A.1 provide more information on the implementation.\footnote{Implementation: \href{https://github.com/bmda-unibas/CondInvarianceCC}{https://github.com/bmda-unibas/CondInvarianceCC}} As an implementation detail, we choose to concatenate the decoded $Z_0$ code with $Z_1$ in order to decode $\hat{X}$, i.e. $\hat{X}=\text{dec}_X (Z_1, \hat{Y}=\text{dec}_Y(Z_0))$. This is an additional measure to ensure that $Z_0$ contains information relevant for property prediction $Y$ and prevent superfluous remaining information about the input $X$ in property space $Z_0$.

%% file: 5-experiments.tex
\section{Experimental Evaluation}
We evaluate the effectiveness of our proposed method w.r.t. (i) selection of a sparse representation with meaningful factors of variation (i.e. model selection) and (ii) enforcing conditional independence in the latent space between these factors. To this end, we conduct experiments on a synthetic dataset with knowledge about appropriate parameterisations to highlight the differences to existing models. Additionally, we evaluate our model on a real-world application with a focus on conditional invariance and generation of novel samples. To assess the performance of our model, we compare our approach to two state-of-the-art baselines: (i) the $\beta$-VAE \cite{Higgins2017betaVAELB} which is a typical baseline model in disentanglement studies and (ii) the symmetry-transformation information bottleneck (STIB) \cite{wieser2020inverse} which ensures conditional invariance through adversarial training and is the direct competitor to our model. We adapt the $\beta$-VAE by adding a decoder for property $Y$ (similar to \cite{Gomez}) which takes only subspace $Z_0$ as input. The latent space of the adapted $\beta$-VAE is split into two subspaces as in the STIB and our model, but has no explicit mechanisms to enforce invariance. This setup can be viewed as an ablation study in which the $\beta$-VAE is the basis model of our approach without cycle consistency and sparsity constraints. The STIB provides an alternative approach for the same goal but with a different mechanism.

The objective of the supervised disentanglement approach is to ensure disentanglement of a fixed property with respect to variations in the invariant space $Z_1$. This is a slightly different setting than in standard unsupervised disentanglement and therefore standard disentanglement metrics might be less insightful. Instead, in order to test the property invariance, we first encode the inputs of the test set and fix the coordinates in the property subspace $Z_0$ which provides prediction $\hat{Y}$. Then we sample uniformly at random in $Z_1$ (plus/minus one standard deviation), decode the generated $\Tilde{X}$ and perform a cycle through the network to obtain $\Tilde{Y}$. This provides the predicted property for the generated $\Tilde{X}$. If conditional invariance between $X$ and $Y$ at a fixed $Z_0$ is warranted, the mean absolute error (MAE) between $\hat{Y}$ and $\Tilde{Y}$ should be close to zero. Thus, all models are 
trained to attain similar MAEs for reconstructing $X$ and, in particular, predicting $Y$, to ensure a fair comparison. 
\subsection{Synthetic Dataset}
In the first experiments, we focus on learning level sets of ellipses and ellipsoids mapped into five dimensions. We consider these experiments as they allow a clear interpretation and visualisation of fixing a property, i.e. choosing the ellipse curve or ellipsoid surface, and known low-dimensional parameterisations are readily available. To this end, we sample uniformly at random data points $X_\text{original}$ from $\mathcal{U}\large([-1,1]^{d_\text{X}}\large)$ and calculate as the corresponding one-dimensional properties $Y_\text{original}$ the ellipse curves ($d_X=2$) and ellipsoid surfaces ($d_X=3$) rotated by $45^\circ$ in the $X_1X_2$-plane. In addition, we add Gaussian noise to the property $Y_\text{original}$. In a real-world scenario, we typically do not have access to the underlying generating process providing $X_\text{original}$ and property $Y_\text{original}$ but a transformed view on these quantities. To reflect this, we map the input $X_\text{original}$ into a five dimensional space ($d'_X=5$), i.e. $X_\text{original}\in[-1,1]^{N\times d_X} \rightarrow X\in\mathbb{R}^{N\times 5}$, and property $Y_\text{original}$ into three dimensional space ($d'_Y=3$), i.e. $Y_\text{original}\in\mathbb{R}_+^{N\times1} \rightarrow Y\in\mathbb{R}^{N\times 3}$, with $N$ data points and dimensions $d_X=\{2,3\}$. See Appendix Sec. A.4 for more details and Fig. A3 for an illustration of the dataset.

Level sets are usually defined implicitly (see Appendix Eq. (A.9)). Common parameterisations consider polar coordinates $(x,y) = (r\cos\varphi, r\sin\varphi)$ for the ellipse and spherical coordinates $(x,y,z)=(r\cos\varphi\sin\theta,r\sin\varphi\sin\theta,r\cos\theta)$ for the ellipsoid, with radius $r\in[0,\infty)$, (azimuth) angle $\varphi\in[0,2\pi)$ in the $X_1X_2$-plane, and polar angle $\theta\in[0,\pi]$ measured from the $X_3$ axis. The goal of our experiment is to identify a low-dimensional parameterisation which captures the underlying radial and angular components, i.e. identify latent dimensions which correspond to parameters $(r,\varphi)$ and $(r,\varphi,\theta)$. 

Details on the architecture and training can be found in Appendix Sec. A.4. We use fully-connected layers for our encoder and decoder networks. Note that in our model, the noise level is fixed at $\sigma_\text{noise}=1$ w.l.o.g. (see Sec. \ref{sec:model_sparseIB}). We choose an 8-dim. latent space, with 3 dimensions reserved for property subspace $Z_0$ and 5 dimensions for invariant subspace $Z_1$. We consider a generous latent space with $d_{Z_1}=d'_X=5$ and $d_{Z_0}=d'_Y=3$ to evaluate the sparsity and model selection. 
\begin{figure}[htbp]
    \centering
    \includegraphics[width=0.89\textwidth]{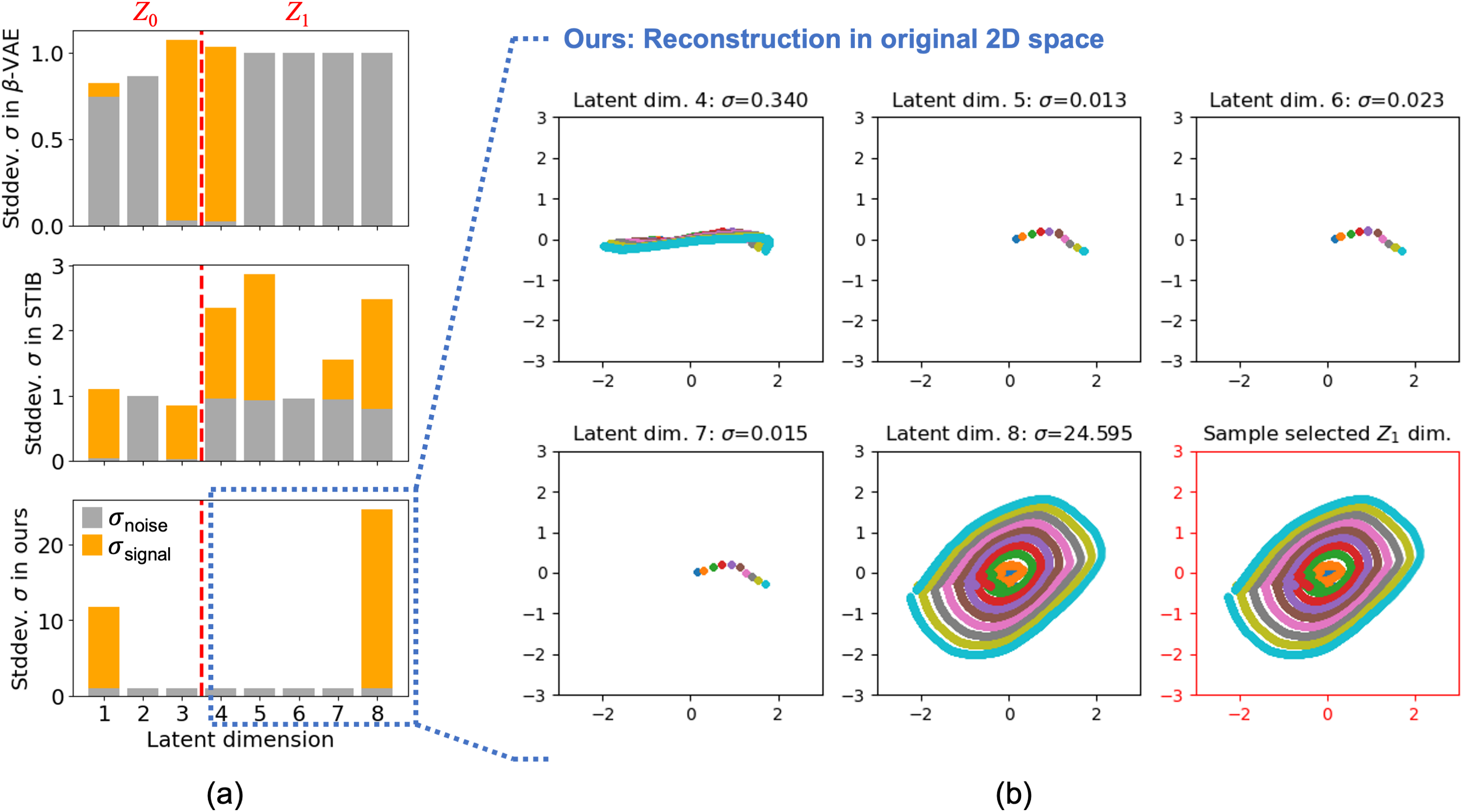} \\
    \vspace{1mm}
    \includegraphics[width=0.89\textwidth]{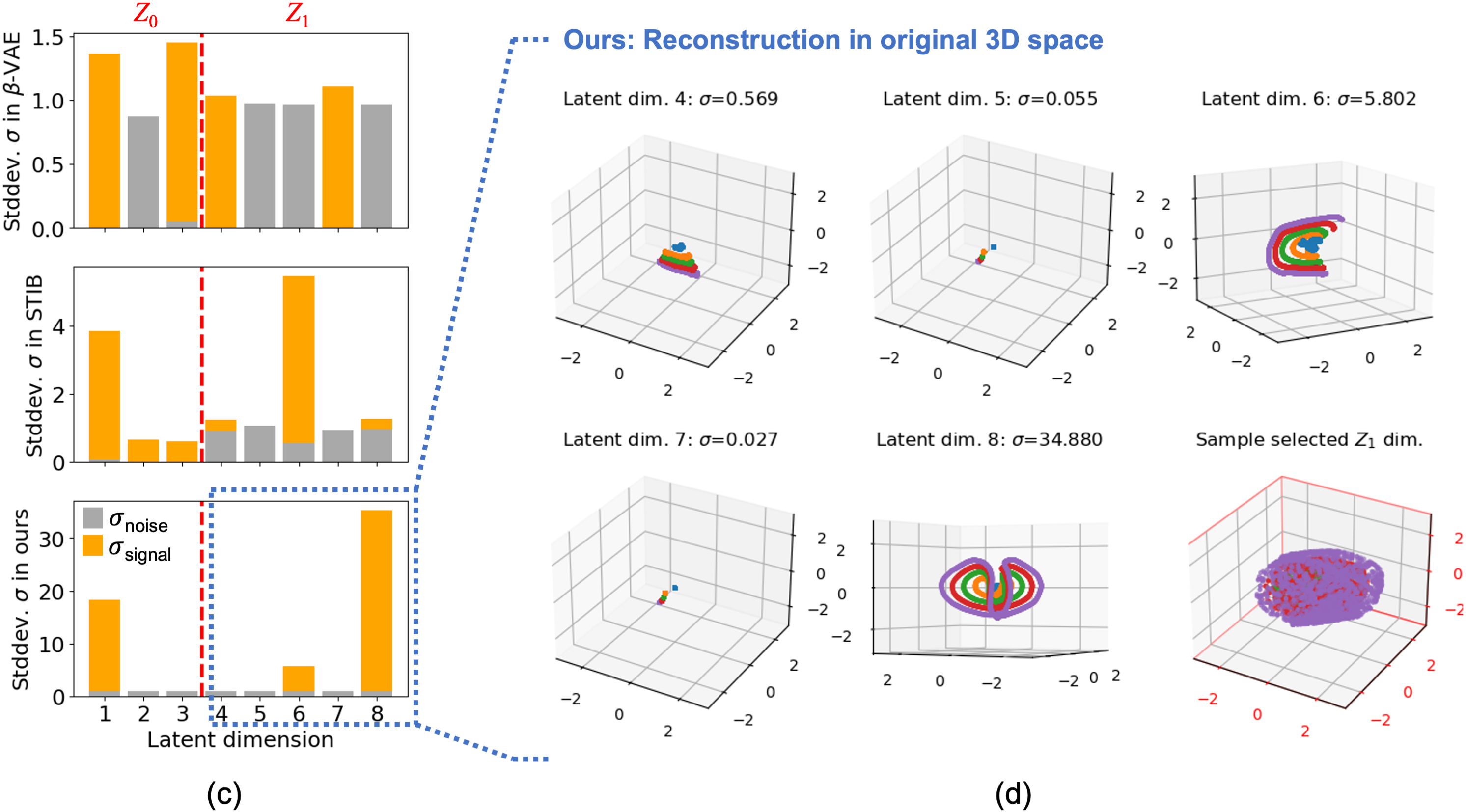}
    \caption{
    Results for ellipse and ellipsoid in original input space ($d_X=\{2,3\}$). (a,c) Illustration of standard deviation in the different latent dimensions, where property subspace $Z_0$ spans dimensions 1-3 and invariant subspace $Z_1$ spans dimensions 4-8. Grey bars indicate the sampling noise $\sigma_\text{noise}$ and orange bars the sample standard deviation $\sigma_\text{signal}$ in the respective dimension. We consider a latent dimension to be selected if the signal exceeds the noise, i.e. orange bars are visible. Only our model selects the expected numbers of parameters. (b,d) Illustration of latent traversal in our model in latent dimensions 4 to 8 in our model in the original input space for fixed values in the property space dimension 1 (different colours). (b) The selected dimension 8 represents the angular component $\varphi$ and reconstructs the full ellipse curves. (d) The selected dimension 6 represents the polar angle $\theta$, while dimension 8 can be related to the azimuth angle $\varphi$. (b,d) The last plot (red borders) samples in all selected dimensions, which reconstructs the full ellipse and ellipsoid, respectively. We intentionally did not sample the ellipsoid surfaces completely to allow seeing surfaces underneath.
    }
    \label{fig:result_synth_2d}
\end{figure}
\subsubsection{Results:}
\label{sec:results_synthetic}
All models attain similar MAEs for $X$ reconstruction and $Y$ prediction but differ in the property invariance as summarised in Table \ref{tab:modelcomp_synthetic}. Our model learns more invariant representations with several factors difference w.r.t. the property invariance in both experiments. In Fig. \ref{fig:result_synth_2d}(a), signal vs. noise for the different models is presented. The standard deviation $\sigma_\text{signal}$ is calculated as the sample standard deviation of the learned means in the respective latent dimension. The sampling noise $\sigma_\text{noise}$ is optimised as a free parameter during training. We consider a latent dimension to be informative or selected if the signal exceeds the noise. The sparest solution is obtained in our model with one latent dimension selected in the property subspace $Z_0$ and one in the invariant subspace $Z_1$. In Fig. \ref{fig:result_synth_2d}(b), we examine the obtained solution more closely in the original data space by mapping back from $d'_X=5$ to $d_X=2$ dimensions. We consider ten equidistant values in the selected $Z_0$ dim. 1 and sample points in the selected $Z_1$ dim. 8. The different colours represent fixed values in $Z_0$, with latent traversal in $Z_1$ dim. 8 reconstructing the full ellipse. This means, the selected latent dim. 8 contains all relevant information at a given coordinate in $Z_0$, while dim. 4 to 7 do not contain any relevant information. We can relate the selected dim. 1 in $Z_0$ to the radius $r$ and dim. 8 in $Z_1$ to the angle $\varphi$. For the ellipsoid ($d_X=3$) we obtain qualitatively the same results as for the ellipse. Again, only our model selects the correct number of latent factors with one in $Z_0$ and two in $Z_1$ (see Fig.\ref{fig:result_synth_2d}(c)). The latent traversal results in Fig. \ref{fig:result_synth_2d}(d) are more intricate to interpret. For latent dim. 6, we obtain a representation which can be interpreted as encoding the polar angle $\theta$. Traversal in latent dim. 8 yields closed curves in three dimensions which can be viewed as on orthogonal representation to dim. 6 and be interpreted as an encoding of the azimuth angle $\varphi$. In both Fig. \ref{fig:result_synth_2d}(b) and \ref{fig:result_synth_2d}(d), the last plot shows sampling in the selected $Z_1$ dimensions for fixed $Z_0$ (i.e. property $Y$) and reconstructs the full ellipse and ellipsoid. Although $\beta$-VAE and STIB perform equally well on reconstructing and predicting on the test set, these models do not consistently lead to sparse and easily interpretable representations which allow direct traversal on the level sets as shown for our model. The presented results remain qualitatively the same for reruns of the models.
\begin{table}[t]
\caption{
Mean absolute errors (MAE) for reconstruction of input $X$, prediction of property $Y$, and property invariance. Ellipse/Ellipsoid: MAEs on 5-dim. input $X$ and 3-dim. property $Y$ are depicted. Molecules: MAEs on input $X$ and property $Y$ as the band gap energy in $\textrm{kcal} \textrm{ mol}^{-1}$.}
\centering
 \begin{tabular}{@{}lcccccccccccc@{}}
 \toprule
 & \multicolumn{3}{c}{Ellipse} & & \multicolumn{3}{c}{Ellipsoid} & & \multicolumn{3}{c}{Molecules} \\ \cmidrule{2-4}\cmidrule{6-8}\cmidrule{10-12}
Model & X & Y & Invar. & & X & Y & Invar. & & X & Y & Invar. \\
\midrule
$\beta$-VAE \hspace{1.5mm}&\hspace{1.5mm} 0.03 \hspace{1.5mm}&\hspace{1.5mm} 0.25  \hspace{1.5mm}&\hspace{1.5mm} 0.058 \hspace{1.5mm}&\hspace{1.5mm} \hspace{1.5mm}&\hspace{1.5mm} 0.02 \hspace{1.5mm}&\hspace{1.5mm} 0.25 \hspace{1.5mm}&\hspace{1.5mm} 0.153 \hspace{1.5mm}&\hspace{1.5mm} \hspace{1.5mm}&\hspace{1.5mm} 0.01 \hspace{1.5mm}&\hspace{1.5mm} 4.01 \hspace{1.5mm}&\hspace{1.5mm} 5.66 \\
STIB \hspace{1.5mm}&\hspace{1.5mm} 0.03 \hspace{1.5mm}&\hspace{1.5mm} 0.25 \hspace{1.5mm}&\hspace{1.5mm} 0.027 \hspace{1.5mm}&\hspace{1.5mm} \hspace{1.5mm}&\hspace{1.5mm} 0.03 \hspace{1.5mm}&\hspace{1.5mm} 0.25 \hspace{1.5mm}&\hspace{1.5mm} 0.083 \hspace{1.5mm}&\hspace{1.5mm} \hspace{1.5mm}&\hspace{1.5mm} 0.01 \hspace{1.5mm}&\hspace{1.5mm} 4.08 \hspace{1.5mm}&\hspace{1.5mm} 3.05\\
\midrule
\textbf{Ours} \hspace{1.5mm}&\hspace{1.5mm} 0.04 \hspace{1.5mm}&\hspace{1.5mm} 0.25 \hspace{1.5mm}&\hspace{1.5mm} \textbf{0.006} \hspace{1.5mm}&\hspace{1.5mm} \hspace{1.5mm}&\hspace{1.5mm} 0.05 \hspace{1.5mm}&\hspace{1.5mm} 0.25 \hspace{1.5mm}&\hspace{1.5mm}  \textbf{0.006} \hspace{1.5mm}&\hspace{1.5mm} \hspace{1.5mm}&\hspace{1.5mm} 0.01 \hspace{1.5mm}&\hspace{1.5mm} 4.06 \hspace{1.5mm}&\hspace{1.5mm}  \textbf{1.34}\\
\bottomrule
\vspace{-5mm}
\end{tabular}
\label{tab:modelcomp_synthetic}
\end{table}
\subsection{Small Organic Molecules (QM9)}
As a more challenging example, we consider the QM9 dataset \cite{rama2014} which includes 133,885 organic molecules. The molecules consist of up to nine heavy atoms (C, O, N, and F), not including hydrogen. Each molecule includes corresponding chemical properties computed with the Density Functional Theory methods. In our experiments, we select a subset with a fixed stoichiometry $(C_7 O_2 H_{10})$ which consists of 6,093 molecules. We choose the band gap energy as the property. 

Details on the architecture and training can be found in Appendix Sec. A.5. We use fully-connected layers for our encoder and decoder. For the input $X$ we use the bag-of-bonds \cite{hansen2015machine} descriptor as a translation, rotation, and permutation invariant representation of molecules, which involves 190 dimensions. 
The latent space size is $17$, where $Z_0$ is 1-dimensional and $Z_1$ is 16-dimensional. 
To evaluate the invariance, we first adjust the regularisation loss weights for a fair comparison of the models. The weights for the irrelevance loss in the STIB and the invariance loss terms in our model were increased until a drop in reconstruction and prediction performances compared to the $\beta$-VAE results was noticeable.
\subsubsection{Results:}
\begin{figure*}[t] 
    \centering
     \hspace{2mm} 
 	\begin{minipage}{.70\textwidth}
 	    \vspace{0mm} 
 		\centering
 		\includegraphics[width=\textwidth]{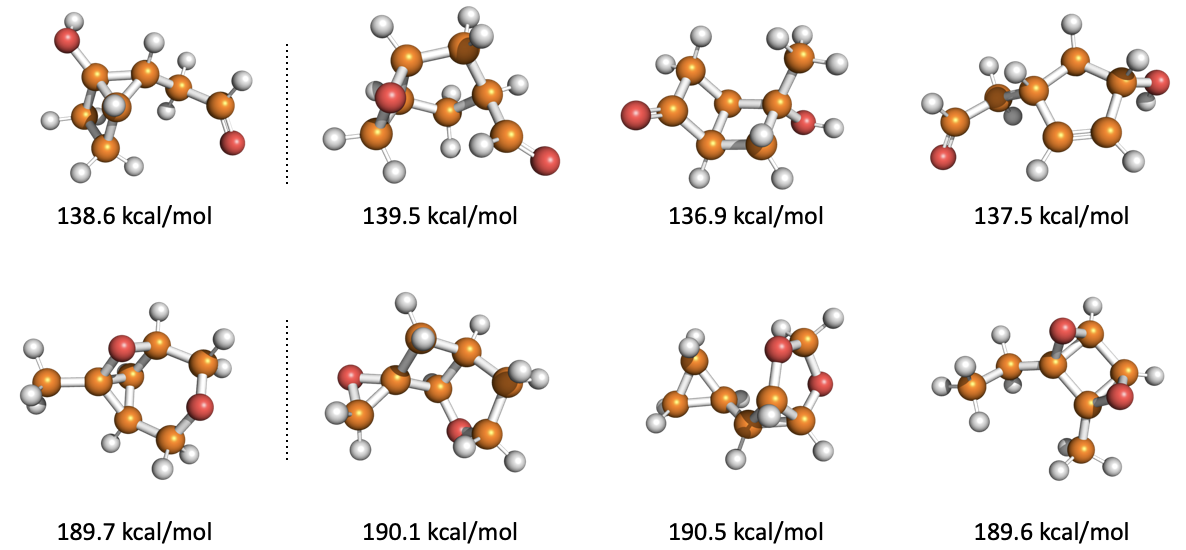}
 		\subcaption{}
		\label{fig:novel_samples}
 	\end{minipage}
     \hspace{8mm}
 	\begin{minipage}{.185\textwidth}
 	    \vspace{0mm}
        \centering
 		\includegraphics[width=\textwidth]{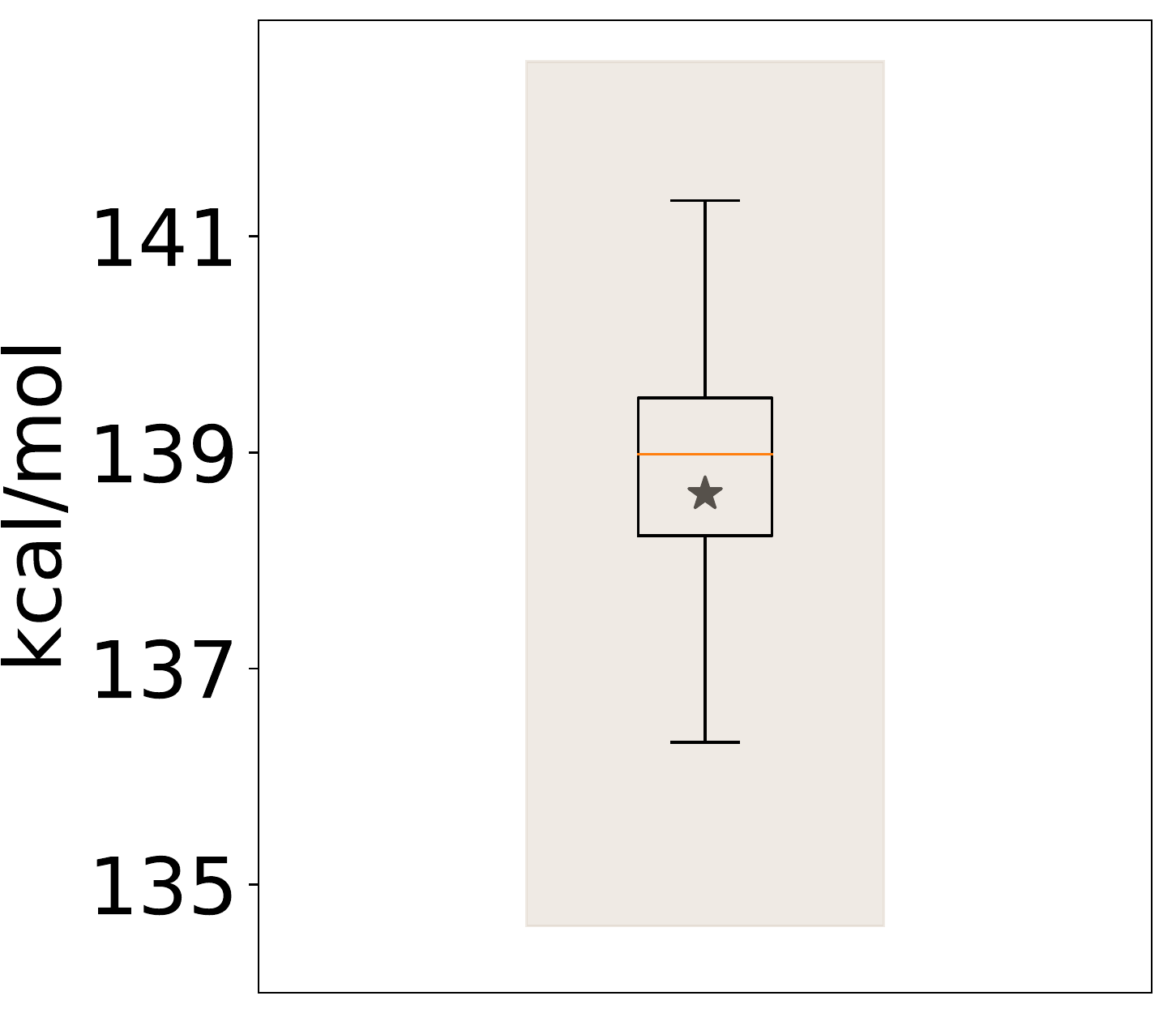}
 		\includegraphics[width=\textwidth]{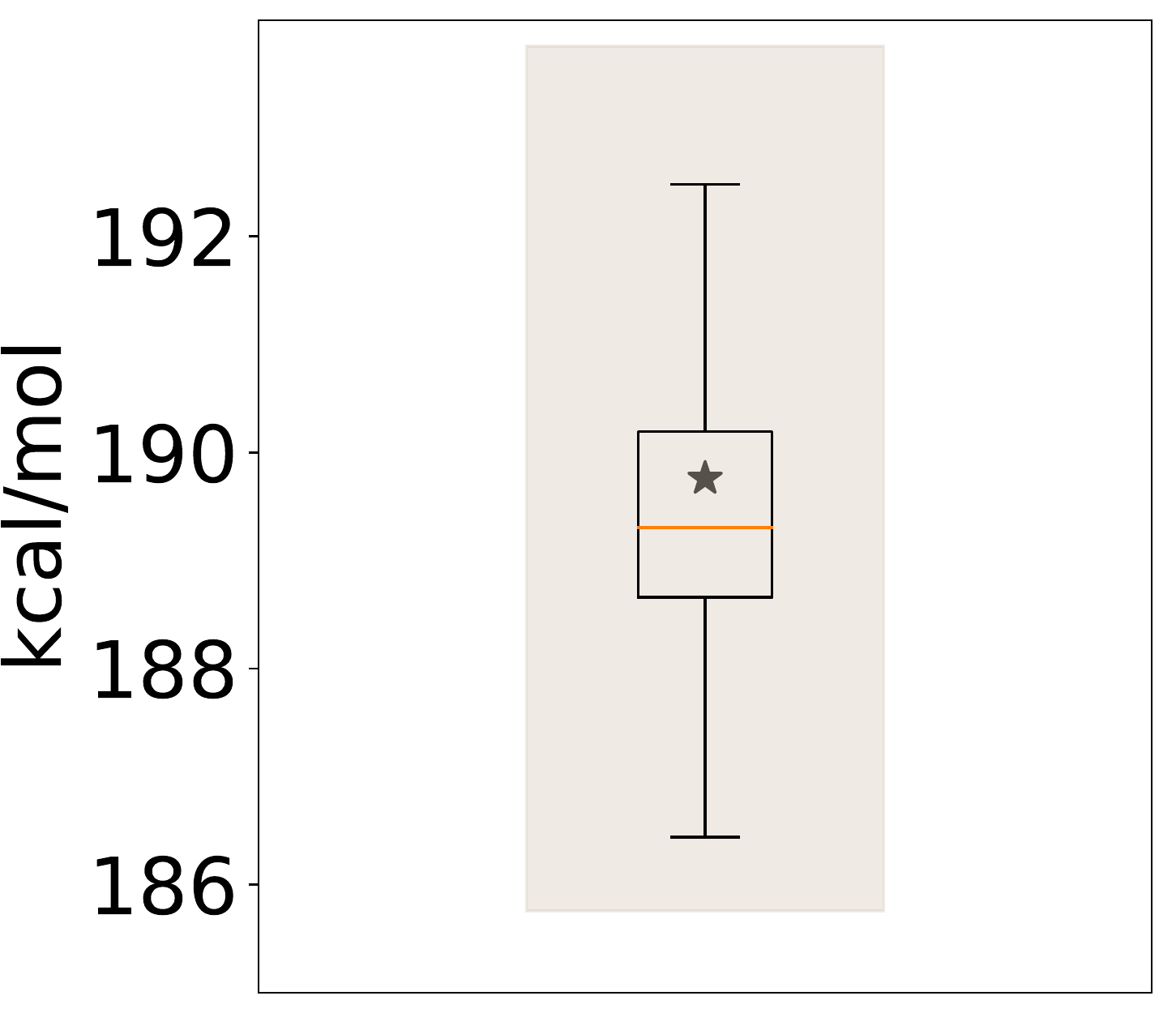}
 		\subcaption{}
 		\label{fig:novel_boxplots}
 	\end{minipage}
 	\caption{Illustration of the generative capability of our model for two reference molecules (rows). (a) The first molecule is the reference molecule with a fixed reference band gap energy. We display three samples and their predicted band gap energies out of 2,000 samples. (b) Boxplots for distribution of predicted property. The star symbol marks the fixed reference band gap energy. The shaded background depicts the prediction error range of the model.}
 	\label{fig:novel}
 \end{figure*}

Table \ref{tab:modelcomp_synthetic} summarises the results. On a test set of 300 molecules, all models achieve similar MAE of 0.01 for the reconstruction of $X$. For prediction of the band gap energies $Y$ a MAE of approx. 4 $\textrm{kcal} \textrm{ mol}^{-1}$ is achieved. The invariance is computed on the basis of 25 test molecules and 400 samples generated for each reference molecule. Similarly to the synthetic experiments, the STIB model performs almost twice as well as the $\beta$-VAE, while our model yields a distinctly better invariance of 1.34 $\textrm{kcal} \textrm{ mol}^{-1}$ among both models. With this result, we can generate novel molecules which are very close to a fixed property. This capability is illustrated in Fig. \ref{fig:novel}. For two reference molecules in the test set, we generate 2,000 new molecules by sampling uniformly at random with one standard deviation in the invariant subspace $Z_1$ and keeping the reference property value, i.e. fixed $Z_0$ coordinates. We show three such examples in Fig. \ref{fig:novel_samples} and select the nearest neighbours in the test set for visualisation of the molecular structure. For all samples, the boxplots in Fig. \ref{fig:novel_boxplots} illustrate the distribution in the predicted property values.
The spread of predicted property values is generally smaller than the model prediction error of 4.06 $\textrm{kcal} \textrm{ mol}^{-1}$ and the predicted property of a majority of samples is close to the target property value.
 

%% file: 6-discussion.tex
\section{Discussion}
\textbf{Sparsity constraints and cycle consistency lead to sparse and interpretable models facilitating model selection.}
The results in Fig. \ref{fig:result_synth_2d}(a,c) demonstrate that our method identifies the sparsest solution in comparison to the standard disentanglement baseline $\beta$-VAE and the direct competitor STIB, which do not address sparsity explicitly. Furthermore, the experiments on ellipses and ellipsoids show that only our model also identifies a correct parameterisation. It correctly learns the radius $r$ in the property subspace $Z_0$ as it encodes the level set, i.e. the ellipse curve or ellipsoid surface given by property $Y$. The angular components $\varphi$ and $\theta$ are correctly -- and in particular independently -- learned in the invariant subspace $Z_1$ (see Fig. \ref{fig:result_synth_2d}(b,d)). This is a direct consequence of the cycle consistency on the property $Y$. It allows for semantically structuring the latent space on the basis of the semantic knowledge on property $Y$. Finally, these results highlight that our method is able to inherently select the correct model. Although the $\beta$-VAE and STIB are capable of attaining similar reconstruction and prediction errors, a reconstruction of level sets in these models requires a more complicated combination of latent dimensions and hinders interpretation. Therefore, only our model makes an interpretation of the learned latent representation feasible.
\newline\newline
\textbf{Cycle consistency enforces conditional invariance.}
Table \ref{tab:modelcomp_synthetic} shows that for all experiments, our model exhibits the best property invariance at otherwise similar reconstruction and prediction errors. The $\beta$-VAE has no mechanisms to ensure invariance and thus performs worst. But although the STIB relies on adversarial training to minimise mutual information (MI) between $Z_1$ and $Y$, the alternating training and MI estimation can pose practical obstacles, especially in cases with high-dimensional latent spaces. Our cycle-consistency-based approach has the same benefits and is more feasible. In particular, our approach can operate on arbitrarily large latent spaces in both $Z_0$ and $Z_1$, because of the inherent sparsity of the solution. Typically, an upper limit for the size of property subspace $Z_0$ and invariant subspace $Z_1$ can be defined by the dimensionality of the property $Y$ and input $X$ (see Fig. \ref{fig:result_synth_2d}). Noteworthy -- although our model is trained and tested on data in the interval $[-1,1]^{d_X}$, $d_X=\{2,3\}$ -- the results generalise well beyond this interval, as long as a part of the level curve or surface was encountered during training (see Fig. \ref{fig:result_synth_2d}(b)). This can be directly attributed to the regularisation of the latent space through additional sampling and cycle consistency of generated samples. These mechanisms impose conditional invariance which, in turn, facilitates generalisation and exploration of new samples by sharing the same level set or symmetry-conserved property.
\newline\newline
\textbf{Conditional invariance improves targeted molecule discovery.}
Conditional invariance is of great importance for the generative potential of our model. In Fig. \ref{fig:novel} we exemplary explored the molecular structures for two reference molecules. By sampling in the invariant space $Z_1$, we discover molecular structures with property values which are very close to the fixed targets, i.e the mean absolute deviation is below the model prediction error. Our experiment demonstrates the ability to generate molecules with self-consistent properties which rely on the improved conditional invariance provided by our model. This facilitates the discovery of novel molecules with desired chemical properties.
\newline
\newline
In conclusion, we demonstrated on synthetic and real-world use cases that our method allows selecting a correct model and improve interpretability as well as exploration of the latent representation. In our synthetic study, we focused on simple cases of connected and convex level sets. To generalise these findings, more general level sets are interesting to be investigated in order to relate to more real-world scenarios. In addition, our approach could be applied to medical applications where a selection of interpretable models is of particular relevance.

%% file: 7-acknowledgements.tex
\subsubsection*{Acknowledgements.} 
This research was supported by the Swiss National Science Foundation through projects No. 167333 within the National Research Programme 75 "Big Data" (M.S.), No. P2BSP2 184359 (S.P.) and the NCCR MARVEL (V.N., M.W., A.W.).

%% file: 121-main.bbl
\begin{thebibliography}{10}
\providecommand{\url}[1]{\texttt{#1}}
\providecommand{\urlprefix}{URL }
\providecommand{\doi}[1]{https://doi.org/#1}

\bibitem{tensorflow2015-whitepaper}
Abadi, M., Agarwal, A., Barham, P., Brevdo, E., Chen, Z., Citro, C., Corrado,
  G.S., Davis, A., Dean, J., Devin, M., Ghemawat, S., Goodfellow, I., Harp, A.,
  Irving, G., Isard, M., Jia, Y., Jozefowicz, R., Kaiser, L., Kudlur, M.,
  Levenberg, J., Man\'{e}, D., Monga, R., Moore, S., Murray, D., Olah, C.,
  Schuster, M., Shlens, J., Steiner, B., Sutskever, I., Talwar, K., Tucker, P.,
  Vanhoucke, V., Vasudevan, V., Vi\'{e}gas, F., Vinyals, O., Warden, P.,
  Wattenberg, M., Wicke, M., Yu, Y., Zheng, X.: {TensorFlow}: Large-scale
  machine learning on heterogeneous systems (2015),
  \url{https://www.tensorflow.org/}, software available from tensorflow.org

\bibitem{achille2018information}
Achille, A., Soatto, S.: Information dropout: Learning optimal representations
  through noisy computation. IEEE Transactions on Pattern Analysis and Machine
  Intelligence  (2018)

\bibitem{ainsworth18a}
Ainsworth, S.K., Foti, N.J., Lee, A.K.C., Fox, E.B.: oi-{VAE}: Output
  interpretable {VAE}s for nonlinear group factor analysis. In: Proceedings of
  the 35th International Conference on Machine Learning (2018)

\bibitem{AlemiFD016}
Alemi, A.A., Fischer, I., Dillon, J.V., Murphy, K.: Deep variational
  information bottleneck. In: 5th International Conference on Learning
  Representations, {ICLR} 2017, Toulon, France, April 24-26, 2017, Conference
  Track Proceedings. OpenReview.net (2017),
  \url{https://openreview.net/forum?id=HyxQzBceg}

\bibitem{multi_level_vae}
Bouchacourt, D., Tomioka, R., Nowozin, S.: Multi-level variational autoencoder:
  Learning disentangled representations from grouped observations. In: AAAI
  Conference on Artificial Intelligence (2018)

\bibitem{art:chechik:gib}
Chechik, G., Globerson, A., Tishby, N., Weiss, Y.: Information bottleneck for
  gaussian variables. In: Journal of Machine Learning Research (2005)

\bibitem{chen2018isolating}
Chen, R.T., Li, X., Grosse, R., Duvenaud, D.: Isolating sources of
  disentanglement in variational autoencoders. arXiv preprint arXiv:1802.04942
  (2018)

\bibitem{chen2016infogan}
Chen, X., Duan, Y., Houthooft, R., Schulman, J., Sutskever, I., Abbeel, P.:
  Infogan: Interpretable representation learning by information maximizing
  generative adversarial nets. arXiv preprint arXiv:1606.03657  (2016)

\bibitem{chicharro2020causal}
Chicharro, D., Besserve, M., Panzeri, S.: Causal learning with sufficient
  statistics: an information bottleneck approach. arXiv preprint
  arXiv:2010.05375  (2020)

\bibitem{Creswell}
Creswell, A., Mohamied, Y., Sengupta, B., Bharath, A.A.: Adversarial
  information factorization (2018)

\bibitem{Gomez}
G{\'o}mez-Bombarelli, R., Wei, J.N., Duvenaud, D., Hern{\'a}ndez-Lobato, J.M.,
  S{\'a}nchez-Lengeling, B., Sheberla, D., Aguilera-Iparraguirre, J., Hirzel,
  T.D., Adams, R.P., Aspuru-Guzik, A.: Automatic chemical design using a
  data-driven continuous representation of molecules. ACS central science
  \textbf{4}(2),  268--276 (2018)

\bibitem{hansen2015machine}
Hansen, K., Biegler, F., Ramakrishnan, R., Pronobis, W., Von~Lilienfeld, O.A.,
  Müller, K.R., Tkatchenko, A.: Machine learning predictions of molecular
  properties: Accurate many-body potentials and nonlocality in chemical space.
  The journal of physical chemistry letters  \textbf{6}(12),  2326--2331 (2015)

\bibitem{Higgins2017betaVAELB}
Higgins, I., Matthey, L., Pal, A., Burgess, C., Glorot, X., Botvinick, M.,
  Mohamed, S., Lerchner, A.: beta-vae: Learning basic visual concepts with a
  constrained variational framework. In: International Conference on Learning
  Representations (2017)

\bibitem{Jha2018DisentanglingFO}
Jha, A.H., Anand, S., Singh, M.K., Veeravasarapu, V.S.R.: Disentangling factors
  of variation with cycle-consistent variational auto-encoders. In: European
  Conference on Computer Vision (2018)

\bibitem{keller2020learning}
Keller, S.M., Samarin, M., Torres, F.A., Wieser, M., Roth, V.: Learning
  extremal representations with deep archetypal analysis. International Journal
  of Computer Vision  \textbf{129}(4),  805--820 (2021)

\bibitem{archetypes}
Keller, S.M., Samarin, M., Wieser, M., Roth, V.: Deep archetypal analysis. In:
  German Conference on Pattern Recognition. pp. 171--185. Springer (2019)

\bibitem{kim2018factorVAE}
Kim, H., Mnih, A.: Disentangling by factorising. In: International Conference
  on Machine Learning. pp. 2649--2658. PMLR (2018)

\bibitem{adam}
Kingma, D.P., Ba, J.: Adam: {A} method for stochastic optimization. In:
  International Conference on Learning Representations (2015)

\bibitem{NIPS2014_5352}
Kingma, D.P., Mohamed, S., Rezende, D.J., Welling, M.: Semi-supervised learning
  with deep generative models. In: Advances in neural information processing
  systems. pp. 3581--3589 (2014)

\bibitem{kingma2013auto}
Kingma, D.P., Welling, M.: Auto-encoding variational bayes. In: Bengio, Y.,
  LeCun, Y. (eds.) 2nd International Conference on Learning Representations,
  {ICLR} 2014, Banff, AB, Canada, April 14-16, 2014, Conference Track
  Proceedings (2014), \url{http://arxiv.org/abs/1312.6114}

\bibitem{Klys}
Klys, J., Snell, J., Zemel, R.: Learning latent subspaces in variational
  autoencoders. In: Advances in Neural Information Processing Systems (2018)

\bibitem{kusner}
Kusner, M.J., Paige, B., Hern{\'a}ndez-Lobato, J.M.: Grammar variational
  autoencoder. In: International Conference on Machine Learning (2017)

\bibitem{FaderNetworks}
Lample, G., Zeghidour, N., Usunier, N., Bordes, A., Denoyer, L., Ranzato, M.:
  Fader networks: Manipulating images by sliding attributes. In: Advances in
  Neural Information Processing Systems (2017)

\bibitem{lin2020infogan}
Lin, Z., Thekumparampil, K., Fanti, G., Oh, S.: Infogan-cr and modelcentrality:
  Self-supervised model training and selection for disentangling gans. In:
  International Conference on Machine Learning. pp. 6127--6139. PMLR (2020)

\bibitem{locatello2019challenging}
Locatello, F., Bauer, S., Lucic, M., Raetsch, G., Gelly, S., Sch{\"o}lkopf, B.,
  Bachem, O.: Challenging common assumptions in the unsupervised learning of
  disentangled representations. In: international conference on machine
  learning. pp. 4114--4124. PMLR (2019)

\bibitem{Louizos}
Louizos, C., Shalit, U., Mooij, J.M., Sontag, D., Zemel, R., Welling, M.:
  Causal effect inference with deep latent-variable models. In: Guyon, I.,
  Luxburg, U.V., Bengio, S., Wallach, H., Fergus, R., Vishwanathan, S.,
  Garnett, R. (eds.) Advances in Neural Information Processing Systems 30, pp.
  6446--6456. Curran Associates, Inc. (2017)

\bibitem{louizos2015variational}
Louizos, C., Swersky, K., Li, Y., Welling, M., Zemel, R.S.: The variational
  fair autoencoder. In: Bengio, Y., LeCun, Y. (eds.) 4th International
  Conference on Learning Representations, {ICLR} 2016, San Juan, Puerto Rico,
  May 2-4, 2016, Conference Track Proceedings (2016),
  \url{http://arxiv.org/abs/1511.00830}

\bibitem{nesterov20203dmolnet}
Nesterov, V., Wieser, M., Roth, V.: 3dmolnet: A generative network for
  molecular structures (2020)

\bibitem{parbhoo20a}
Parbhoo, S., Wieser, M., Roth, V., Doshi-Velez, F.: Transfer learning from
  well-curated to less-resourced populations with hiv. In: Proceedings of the
  5th Machine Learning for Healthcare Conference (2020)

\bibitem{parbhoo2020information}
Parbhoo, S., Wieser, M., Wieczorek, A., Roth, V.: Information bottleneck for
  estimating treatment effects with systematically missing covariates. Entropy
  \textbf{22}(4), ~389 (2020)

\bibitem{rama2014}
Ramakrishnan, R., Dral, P.O., Rupp, M., Von~Lilienfeld, O.A.: Quantum chemistry
  structures and properties of 134 kilo molecules. Scientific data
  \textbf{1}(1), ~1--7 (2014)

\bibitem{sudhir}
Raman, S., Fuchs, T.J., Wild, P.J., Dahl, E., Roth, V.: The bayesian
  group-lasso for analyzing contingency tables. In: Proceedings of the 26th
  Annual International Conference on Machine Learning (2009)

\bibitem{rey14}
Rey, M., Roth, V., Fuchs, T.: Sparse meta-gaussian information bottleneck. In:
  International Conference on Machine Learning. pp. 910--918. PMLR (2014)

\bibitem{rezende14}
Rezende, D.J., Mohamed, S., Wierstra, D.: Stochastic backpropagation and
  approximate inference in deep generative models. In: International Conference
  on Machine Learning (2014)

\bibitem{robert2019dualdis}
Robert, T., Thome, N., Cord, M.: Dualdis: Dual-branch disentangling with
  adversarial learning (2019)

\bibitem{Simon13asparse-group}
Simon, N., Friedman, J., Hastie, T., Tibshirani, R.: A sparse-group lasso.
  JOURNAL OF COMPUTATIONAL AND GRAPHICAL STATISTICS  (2013)

\bibitem{song2019understanding}
Song, J., Ermon, S.: Understanding the limitations of variational mutual
  information estimators. arXiv preprint arXiv:1910.06222  (2019)

\bibitem{tibshirani96regression}
Tibshirani, R.: Regression shrinkage and selection via the lasso. Journal of
  the Royal Statistical Society (Series B)  (1996)

\bibitem{art:tishby:ib}
Tishby, N., Pereira, F.C., Bialek, W.: The information bottleneck method. In:
  Allerton Conference on Communication, Control and Computing (1999)

\bibitem{wieczorek2016causal}
Wieczorek, A., Roth, V.: Causal compression. arXiv preprint arXiv:1611.00261
  (2016)

\bibitem{alex_bound}
Wieczorek, A., Roth, V.: On the difference between the information bottleneck
  and the deep information bottleneck. Entropy  \textbf{22}(2), ~131 (2020)

\bibitem{Wieczorek}
Wieczorek, A., Wieser, M., Murezzan, D., Roth, V.: Learning sparse latent
  representations with the deep copula information bottleneck. In: 6th
  International Conference on Learning Representations, {ICLR} 2018, Vancouver,
  BC, Canada, April 30 - May 3, 2018, Conference Track Proceedings.
  OpenReview.net (2018), \url{https://openreview.net/forum?id=Hk0wHx-RW}

\bibitem{mariophd}
Wieser, M.: Learning Invariant Representations for Deep Latent Variable Models.
  Ph.D. thesis, University\_of\_Basel (2020)

\bibitem{wieser2020inverse}
Wieser, M., Parbhoo, S., Wieczorek, A., Roth, V.: Inverse learning of
  symmetries. In: Advances in Neural Information Processing Systems (2020)

\bibitem{wu2017sparsity}
Wu, M., Hughes, M.C., Parbhoo, S., Zazzi, M., Roth, V., Doshi-Velez, F.: Beyond
  sparsity: Tree regularization of deep models for interpretability (2017)

\bibitem{cyclegan}
{Zhu}, J., {Park}, T., {Isola}, P., {Efros}, A.A.: Unpaired image-to-image
  translation using cycle-consistent adversarial networks. In: International
  Conference on Computer Vision (2017)

\end{thebibliography}
